\documentclass[sigconf,screen]{acmart}

\usepackage{microtype}
\usepackage{subfigure}
\usepackage{multirow} 
\usepackage{booktabs} 
\usepackage[utf8]{inputenc}
\usepackage[dvipsnames]{xcolor}
\usepackage[most]{tcolorbox}

\usepackage[table]{xcolor}
\newcommand{\topk}{\mathrm{top}\text{-}k}
\AtBeginDocument{%
  }

\copyrightyear{2026}
\acmYear{2026}
\setcopyright{cc}
\setcctype{by}
\acmConference[ICMR '26]{International Conference on Multimedia Retrieval}{June 16--19, 2026}{Amsterdam, Netherlands}
\acmBooktitle{International Conference on Multimedia Retrieval (ICMR '26), June 16--19, 2026, Amsterdam, Netherlands}
\acmDOI{10.1145/3805622.3810730}
\acmISBN{979-8-4007-2617-0/2026/06}

\begin{document}

\title{A-MAR: Agent-based Multimodal Art Retrieval for Fine-Grained Artwork Understanding}

\author{Shuai Wang}
\email{s.wang3@uva.nl}
\affiliation{%
  \institution{University of Amsterdam}
  \city{Amsterdam}
  \country{NL}
}
\authornote{Part of this work was done during interning at Amazon.}

\author{Hongyi Zhu}
\email{h.zhu@uva.nl}
\affiliation{%
  \institution{University of Amsterdam}
  \city{Amsterdam}
  \country{NL}
}

\author{Jia-Hong Huang}
\email{j.huang@uva.nl}
\affiliation{%
  \institution{University of Amsterdam}
  \city{Amsterdam}
  \country{NL}
}
\additionalaffiliation{%
\institution{Amazon AGI}
}

\author{Yixian Shen}
\email{y.shen@uva.nl}
\affiliation{%
  \institution{University of Amsterdam}
  \city{Amsterdam}
  \country{NL}
}

\author{Chengxi Zeng}
\email{simon.zeng@bristol.ac.uk}
\affiliation{%
  \institution{University of Bristol}
  \city{Bristol}
  \country{United Kingdom}
}

\author{Stevan Rudinac}
\email{s.rudinac@uva.nl}
\affiliation{%
  \institution{University of Amsterdam}
  \city{Amsterdam}
  \country{NL}
}

\author{Monika Kackovic}
\email{m.kackovic@uva.nl}
\affiliation{%
  \institution{University of Amsterdam}
  \city{Amsterdam}
  \country{NL}
}

\author{Nachoem Wijnberg}
\email{n.m.wijnberg@uva.nl}
\affiliation{%
  \institution{University of Amsterdam}
  \city{Amsterdam}
  \country{NL}
}
\additionalaffiliation{%
\institution{College of Business and Economics,
University of Johannesburg}
  \city{Johannesburg}
  \country{South Africa}
  }

\author{Marcel Worring}
\email{m.worring@uva.nl}
\affiliation{%
  \institution{University of Amsterdam}
  \city{Amsterdam}
  \country{NL}
}

\renewcommand{\shortauthors}{Shuai Wang et al.}

\renewcommand{\shortauthors}{Shuai Wang et al.}


\begin{abstract}
Understanding artworks requires multi-step reasoning over visual content and cultural, historical, and stylistic context. While recent multimodal large language models show promise in artwork explanation, they rely on implicit reasoning and internalized knowledge, limiting interpretability and explicit evidence grounding. We propose \textbf{A-MAR}, an \textbf{A}gent-based \textbf{M}ultimodal \textbf{A}rt \textbf{R}etrieval framework that explicitly conditions retrieval on structured reasoning plans.
Given an artwork and a user query, A-MAR first decomposes the task into a structured reasoning plan that specifies the goals and evidence requirements for each step. Retrieval is then conditioned on this plan, enabling targeted evidence selection and supporting step-wise, grounded explanations.
To evaluate agent-based multimodal reasoning within the art domain, we introduce ArtCoT-QA. This diagnostic benchmark features multi-step reasoning chains for diverse art-related queries, enabling a granular analysis that extends beyond simple final answer accuracy. 
Experiments on SemArt and Artpedia show that A-MAR consistently outperforms static, non-planned retrieval and strong MLLM baselines in final explanation quality, while evaluations on ArtCoT-QA further demonstrate its advantages in evidence grounding and multi-step reasoning ability. These results highlight the importance of reasoning-conditioned retrieval for knowledge-intensive multimodal understanding and position A-MAR as a step toward interpretable, goal-driven AI systems, with particular relevance to cultural industries. The code and data are available at: \url{https://github.com/ShuaiWang97/A-MAR}.

\end{abstract}

\ccsdesc[500]{Applied computing~Fine arts}
\ccsdesc[500]{Computing methodologies~Natural language generation}

\keywords{Artwork Analysis, Multimodal reasoning, Multimodal retrieval}

\maketitle

\section{Introduction}

\begin{figure}[!t]
	\centering
	\includegraphics[width=\columnwidth, trim=0.7cm 0cm 1cm 0cm,
    clip]{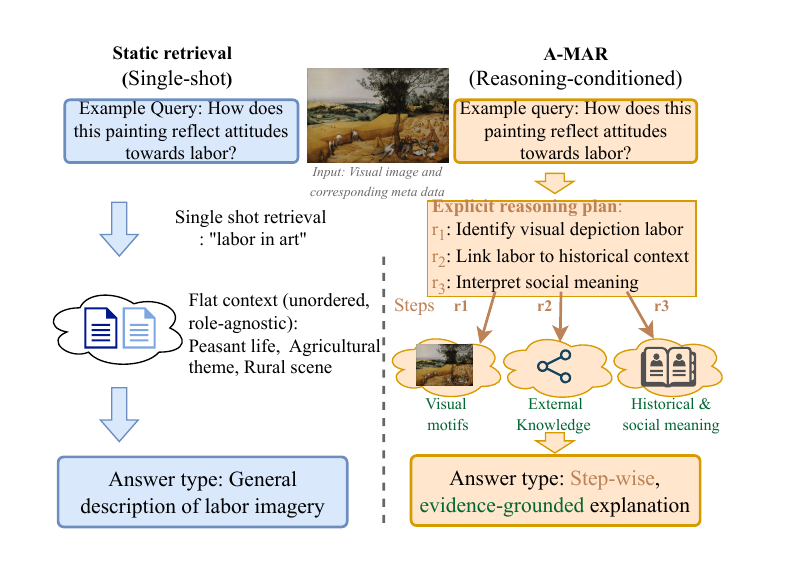}\vspace{-4mm}
	\caption{Comparison between static retrieval and A-MAR for open-ended, user-specified artwork question. While static approach retrieves a flat, unordered context in response to the query, A-MAR explicitly plans the reasoning process and retrieves step-specific evidence across modalities, enabling interpretable and grounded explanations. }
    \vspace{-4mm}
    \label{fig:example}
\end{figure}

Understanding fine art is a fundamentally multimodal and reasoning-intensive task that goes beyond surface-level image description. Meaningful interpretation requires not only recognizing visual elements, but also reasoning over symbolic motifs, stylistic conventions, and cultural–historical contexts that are often not directly observable in the image~\cite{BMVC_style_2013,panofsky1955meaning, baxandall1988painting}. For example, interpreting a Renaissance painting may involve identifying visual motifs, linking them to religious or philosophical symbolism, and positioning them within broader historical narratives.  Crucially, such interpretation is an ordered, multi-step reasoning process in which different steps require different types of evidence, rather than a single-pass synthesis of information, as illustrated in Figure~\ref{fig:example}.

Recent multimodal large language models (MLLMs) have demonstrated strong performance on general image captioning and visual question answering tasks~\cite{NEURIPS2023_llava, chen2023sharegpt4v, qwen2vl, yu2025crispsam2sam2crossmodalinteraction, zhu2024enhancing}. However, in the domain of fine art, these models often struggle to produce reliable and interpretable explanations, as their reasoning remains implicit and relies on internalized knowledge that may be incomplete or hallucinated~\cite{llm_Hallucination_2023, VLM_Hallucination_2024, rudin2019stop}. To mitigate these limitations, prior work has explored retrieval-augmented generation (RAG)~\cite{zhao2025surveylargelanguagemodels,brown2020,rag_review}, which incorporates external knowledge at inference time to improve factual grounding. Despite its promise, most RAG-based systems adopt static, single-shot retrieval strategies that return a flat, unordered set of context documents~\cite{izacard-grave-2021-leveraging, rudin2019stop}. Such query-driven retrieval assumes all evidence plays an equal role, failing to account for the fact that different reasoning steps require different types of evidence drawn from different modalities~\cite{yao2023react, zhao2023retrievingmultimodalinformationaugmented}. Building on this line of work, recent approaches in the art domain~\cite{ijcai2024p848,ArtRAG_MM2025,lee-etal-2024-multimodal} have shown that incorporating structured contextual knowledge can significantly improve multimodal artwork understanding and explanation. While these approaches advance beyond unstructured text retrieval, they still rely on static, query-driven retrieval pipelines that use the user question directly to retrieve context~\cite{yao2023react,Singh2025AgenticRG}. As a result, retrieval remains agnostic of the internal structure of the reasoning process: the system does not explicitly determine what information is needed and when it should be retrieved for each reasoning step. This limitation becomes particularly pronounced for complex questions requiring multi-step interpretation, where visual evidence, historical background, and symbolic meaning must be integrated in a specific order. Consequently, existing systems lack explicit and controllable mechanisms for reasoning-conditioned retrieval, leaving a critical gap between structured knowledge and effective multimodal reasoning for fine-grained artwork understanding.

Inspired by established interpretive practices in art history~\cite{panofsky1955meaning,baxandall1988painting}, in which analysis proceeds iteratively from visual observation to contextualization and interpretation, we introduce Agent-based Multimodal Art Retrieval (A-MAR).
Unlike prior RAG-based approaches~\cite{ijcai2024p848, ArtRAG_MM2025, lee-etal-2024-multimodal}, A-MAR introduces an explicit reasoning and planning phase that decomposes a complex art-related question into an ordered sequence of reasoning steps, each associated with specific evidence requirements. Rather than relying on a static retrieval strategy, the system autonomously determines what relevant contextual knowledge is needed and when it should be retrieved before generating the final explanation. By aligning retrieval behavior with explicit reasoning plans, A-MAR enables controlled, interpretable, and evidence-grounded multimodal reasoning suited to the complexity of fine-grained art understanding.

To rigorously evaluate agent-based multimodal reasoning in fine-grained art understanding, we introduce ArtCoT-QA, a diagnostic benchmark of artwork-centered questions annotated with explicit multi-step reasoning chains and fine-grained evidence grounding. Unlike existing art explanation datasets that provide only final descriptions, ArtCoT-QA enables step-level analysis of whether a model’s reasoning process is interpretable and faithfully grounded in appropriate evidence.  On the reasoning-intensive ArtCoT-QA benchmark, A-MAR yields up to $15\%$ higher performance in step completeness, reasoning faithfulness, and answer quality compared to static retrieval and text-only planning baselines. On SemArt and Artpedia, A-MAR improves final artwork explanation quality over existing approaches for visual art understanding, achieving gains of up to $+3.9$ BLEU and $+1.9$ SPICE under the same backbone model. These results position A-MAR as a strong foundation for interactive and explainable multimodal reasoning systems for cultural industry applications.
\begin{itemize}

  \item We propose \textbf{A-MAR}, an agent-based multimodal retrieval framework for multimodal art reasoning, which introduces an explicit planning phase to jointly structure retrieval and reasoning over knowledge sources.

 \item We introduce \textbf{ArtCoT-QA}, a benchmark for complex artwork question answering that supports step-level reasoning evaluation and multimodal evidence grounding, addressing limitations of existing art explanation datasets.

 \item  We conduct extensive experiments and ablation studies across multiple datasets, demonstrating that multimodal planning improves retrieval relevance, evidence grounding, and multi-step reasoning quality in multimodal art understanding.

\end{itemize}

\section{Related work}
We discuss related work on two key relevant subjects: multimodal art understanding and reasoning-aware multimodal retrieval.

\subsection{Multimodal art understanding}

Recent multimodal large language models (MLLMs) have demonstrated strong performance on image captioning, visual question answering, and general multimodal reasoning by integrating visual encoders with large language models \cite{chen2023sharegpt4v, clip_2021,2023_BLIP, NEURIPS2023_llava}. In the visual art domain, prior work has explored painting classification~\cite{strezoski2020tindart, Art_ICIP_16, eccv12_art, 2017omniart, proto_HGNN}, retrieval~\cite{conde_clip-art_2021, BMVC_art_retrieval, Thanos_MM_21, semart_2018}, captioning~\cite{MM_19_art, Iconographic_IC, achlioptas2021artemis, mohamed_it_2022, li2021paint4poemdatasetartisticvisualization}, and fine-tuning MLLMs for art-specific analysis such as formal, emotional, and general interpretation~\cite{MM24GalleryGPT, Yuan2023ArtGPT4TA, ExpArt_acl2024, efthymiou2026vl}.
Despite this progress, visual art understanding requires reasoning beyond surface-level visual perception, involving symbolic meaning, artistic style, and historical and cultural context that are often not directly observable in the image. While recent approaches incorporate contextual knowledge by structuring explanations \cite{Bai2021ExplainMT} or injecting structured metadata via heterogeneous graphs \cite{ijcai2024p848}, they rely on unstructured retrieval or predefined schemas. As a result, reasoning remains largely implicit and weakly aligned with supporting evidence, limiting interpretability and control. This motivates approaches that explicitly structure both contextual knowledge and the reasoning process for fine-grained artwork understanding.

\subsection{Reasoning-aware multimodal retrieval}

Reasoning-aware retrieval plays a central role in grounding large language models by selecting external evidence that supports interpretation and multi-step reasoning \cite{Zhao2023RetrievingMI, rag_survey, edge2025localglobalgraphrag, guo2024lightragsimplefastretrievalaugmented, wang2024ada, zhu_tomm_2025}. Most existing retrieval augmented  systems adopt static, single-shot retrieval strategies, where a static query directly determines the retrieved context, limiting their ability to support multi-step reasoning or adapt retrieval to different reasoning needs.
In the art domain, methods such as KALE \cite{ijcai2024p848} and ArtRAG \cite{ArtRAG_MM2025} demonstrate the benefits of integrating structured knowledge, including artists, styles, and historical context, into multimodal generation. However, retrieval in these systems remains static and over-reliant on the initial query, offering limited control over how evidence is selected or adapted across different stages of interpretation.
Recent agent and planning-based RAG approaches introduce explicit reasoning or query planning to guide retrieval decisions \cite{Singh2025AgenticRG, rag_review, Modular_Gen_IR_2025, khan2024exquisitor, khan2024exquisitor2}. Nevertheless, such works primarily focus on text-centric tasks and do not address the challenges of multimodal, domain-specific reasoning. Fine-grained artwork understanding requires coordinated reasoning over visual content, metadata, and structured cultural knowledge. A-MAR addresses this gap by introducing explicit multimodal planning into structured art knowledge retrieval, enabling controlled and evidence-grounded reasoning tailored to the art domain.

\begin{figure*}[ht]
    \centering\includegraphics[width=\linewidth]{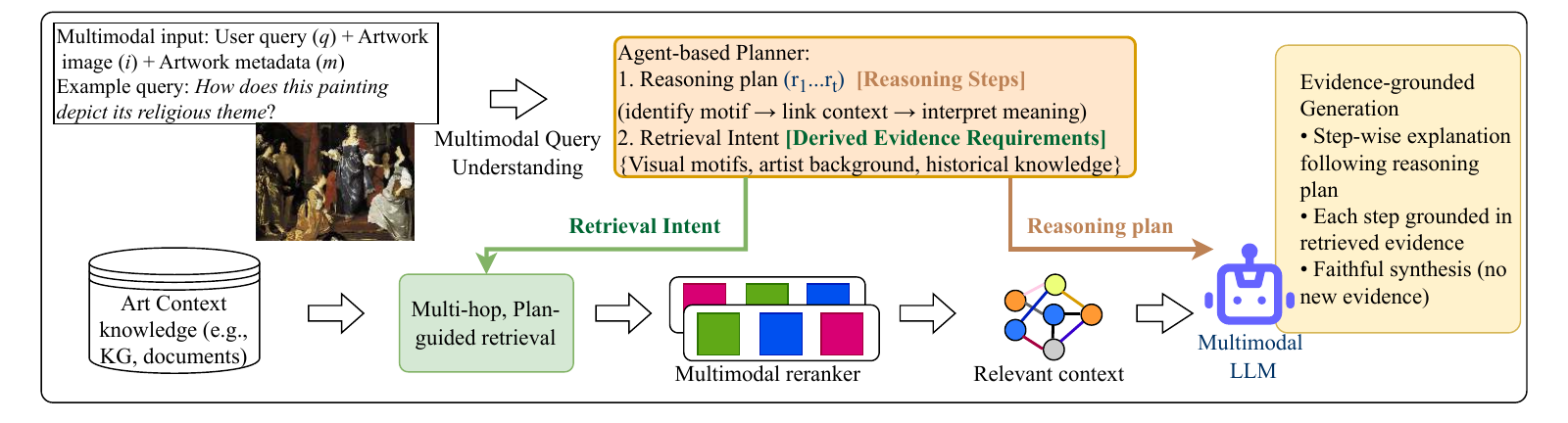}
    \vspace{-5mm}
    \caption{Overview of A-MAR framework for agent-based multimodal artwork reasoning. Given a multimodal artwork query, an agentic planner produces an explicit reasoning plan and retrieval intent, which guide multi-hop retrieval over an art context knowledge base. Retrieved evidence is re-ranked and assembled into a compact context, and a multimodal LLM generates a step-wise explanation by following the reasoning plan and grounding each step in the retrieved evidence.}
    \label{fig:icmr2026_framework_2_1}
\end{figure*}

\section{Method: Agent-based multimodal art retrieval }
To this end, we propose A-MAR, an agent-based retrieval augmented framework that explicitly separates \emph{reasoning}, \emph{retrieval}, and \emph{generation}. As illustrated in Figure~\ref{fig:icmr2026_framework_2_1}, A-MAR first externalizes reasoning as an explicit, structured plan that decomposes the question $Q$ into an ordered sequence of reasoning steps, each associated with a specific evidence requirement. This plan serves as a control interface that governs how external knowledge is retrieved and how evidence is used during answer generation.

\subsection{Problem setup and overview}

We study the problem of \emph{Question-conditioned multimodal artwork understanding}, where the goal is to generate not only accurate but interpretable and evidence-grounded answers. Given an artwork image $I$, associated metadata $M$ (e.g., artist, title, timeframe, technique), and a user question $Q$, the goal is to generate an answer $A$ that is not only accurate, but also interpretable and grounded in appropriate visual and contextual evidence.

Unlike conventional RAG systems that perform retrieval directly from the user query, A-MAR explicitly models reasoning as a structured planning process that guides retrieval. The framework decomposes inference into three stages:
(1) \emph{Reasoning and planning} stage produces a reasoning plan $\mathcal{P}_{\text{gen}} = \{(r_t, e_t)\}_{t=1}^T$, where each step $r_t$ specifies a sub-goal of the reasoning process and $e_t$ denotes the type of evidence required (e.g., visual, metadata, or structured knowledge).
(2) A \emph{reasoning-conditioned retrieval} stage retrieves a compact related information from external art knowledge documents based on the planner’s retrieval intent, rather than directly using the user question as a static query. 
 (3) An \emph{evidence-grounded generation} stage synthesizes the final answer $A$ by following the reasoning plan step-by-step and grounding each step in the retrieved evidence and visual input.  A-MAR is designed to study how explicit planning influences retrieval behavior and reasoning faithfulness in multimodal art understanding. It focuses on structuring and controlling \emph{what evidence is retrieved and why}, enabling interpretable, goal-driven reasoning for complex artwork questions.

\begin{figure*}[ht]
    \centering\includegraphics[width=\linewidth]{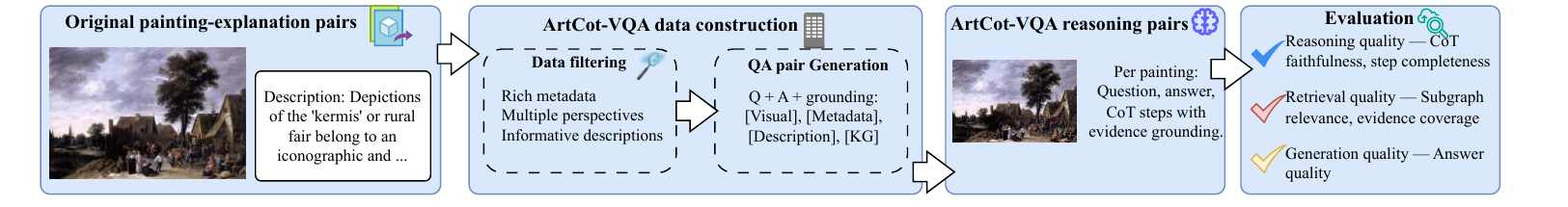}
    \vspace{-3mm}
    \caption{ArtCoT-QA dataset construction and evaluation protocol. Starting from curated artwork images, metadata, and textual descriptions, ArtCoT-QA constructs reasoning-intensive question–answer pairs annotated with multi-step chain-of-thought and step-level evidence grounding. The benchmark enables fine-grained evaluation of reasoning, retrieval, and generation quality in multimodal artwork question answering.}
    \label{fig:icmr2026_framework_2_2}
    \vspace{-3mm}

\end{figure*}

\subsection{Agent-based reasoning and planning}

The agent-inspired reasoning and planning module serves as the core control component of A-MAR. Rather than generating answers, it explicitly specifies the structure of reasoning by determining the ordered sub-goals required to answer a query and the type of evidence needed to support each step. By externalizing this process, A-MAR separates reasoning structure from retrieval and generation, enabling interpretable multimodal inference.

Formally, the planner operates on the artwork image $I$, associated metadata $M$, and the user question $Q$, and is implemented using a vision--language model. It produces a structured reasoning plan
\begin{equation}
\mathcal{P}_{\text{gen}} = \{(r_t, e_t)\}_{t=1}^T,
\end{equation}
where each $r_t$ denotes a reasoning sub-goal and $e_t \in \mathcal{E}$ specifies the primary evidence type required at step $t$ (e.g., visual content, metadata, or contextual knowledge). The planner can be viewed as a mapping
\begin{equation}
\Phi_{\text{plan}} : (I, M, Q) \rightarrow \mathcal{P}_{\text{gen}},
\end{equation}
which defines the intended reasoning flow without committing to intermediate conclusions.

From the reasoning plan, the planner derives a retrieval intent that aggregates step-level evidence requirements across $\mathcal{P}_{\text{gen}}$. This intent is expressed as a retrieval query $q_{\text{ret}}$, obtained via
\begin{equation}
q_{\text{ret}} = \psi(\mathcal{P}_{\text{gen}}),
\end{equation}
where $\psi(\cdot)$ abstracts \emph{what evidence is required} from \emph{how the question is phrased}. Importantly, $q_{\text{ret}}$ is not a paraphrased $Q$, but a structured representation of the planner’s evidence needs, serving as the interface between reasoning and retrieval.

Crucially, the planner performs \emph{meta-reasoning} rather than answer generation. The planner produces a machine-readable plan in a restricted output space $(r_t, e_t)$, avoiding premature content generation and enabling explicit analysis of reasoning structure and evidence alignment.
The agent-based nature of the planner lies in this goal-directed formulation: it autonomously decomposes complex, open-ended queries into interpretable reasoning sub-goals and evidence requirements that subsequently control retrieval and generation. By treating reasoning as an explicit control plan rather than a static retrieval, A-MAR enables principled reasoning-conditioned retrieval and faithful, step-wise multimodal explanation.

\subsection{Reasoning-conditioned retrieval}

Given the planner outputs, A-MAR performs \emph{reasoning-conditioned retrieval}, where evidence selection is guided by a retrieval intent $q_{\text{ret}}$ derived from the reasoning plan $\mathcal{P}_{\text{gen}}=\{(r_t,e_t)\}_{t=1}^T$, rather than directly from the user question $Q$. The retrieval intent abstracts the planner’s assessment of what evidence is required across all reasoning steps.
Retrieval proceeds in two stages.  

\textbf{(1) Coarse retrieval.} We embed the retrieval intent $q_{\text{ret}}$ and candidate knowledge units (e.g., entity descriptions or text passages) into a shared representation space using an encoder $f(\cdot)$. A candidate set is obtained by semantic similarity $\mathrm{sim}(.,.)$ between representation in shared emebdding space:
\begin{equation}
\mathcal{C}_k = \arg\topk_{c} \ \mathrm{sim}\big(f(q_{\text{ret}}), f(c)\big).
\end{equation}

\textbf{(2) Reranking.} The candidate set is refined by a joint scoring function that measures semantic alignment with the retrieval intent and the artwork image $I$, using a vision--language model scorer $g$:
\begin{equation}
s_{\text{sem}}(c) = g(I, q_{\text{ret}}, c).
\end{equation}
When the underlying knowledge source has relational structure (e.g., a knowledge graph), an additional structural importance term $s_{\text{str}}(c)$ (such as node degree or centrality) can be incorporated. The final score is
\begin{equation}
s(c) = \lambda \cdot s_{\text{sem}}(c) + (1-\lambda)\cdot s_{\text{str}}(c),
\end{equation}
and the top-ranked candidates are retained as the retrieved context.
Although the reasoning plan consists of multiple ordered steps, retrieval is performed once using $q_{\text{ret}}$, ensuring coverage of step-level evidence requirements without increasing retrieval frequency. The retrieved context forms a shared evidence pool, which the generator subsequently interprets according to $\mathcal{P}_{\text{gen}}$, decoupling \emph{evidence selection} from \emph{evidence utilization}.

\subsection{Evidence-grounded generation}

The final stage of A-MAR generates the answer by explicitly grounding generation in both retrieved evidence and the reasoning plan. Given the user question $Q$, artwork image $I$, metadata $M$, retrieved context subgraph, and generation plan $\mathcal{P}_{\text{gen}} = \{(r_t, e_t)\}_{t=1}^T$, the generator produces an answer $A$ by following the plan sequentially and instantiating each reasoning step with its required evidence.

Unlike conventional RAG systems that concatenate retrieved context into a single prompt, A-MAR formulates generation as a plan-constrained process. The reasoning plan specifies both the reasoning order and the evidence type for each step, and the generator is instructed to (i) follow the reasoning sequence in $\mathcal{P}_{\text{gen}}$ and (ii) ground each step in evidence consistent with its evidence requirement $e_t$.
This design enforces two key properties.  
\emph{Evidence alignment:} each reasoning step is explicitly associated with an evidence type (e.g., visual cues, metadata, or structured contextual knowledge), enabling traceability between generated content and its supporting evidence.  
\emph{Separation of concerns:} evidence selection is handled by the planning and retrieval stages, while generation focuses solely on composing a coherent explanation from the provided inputs, isolating the effect of reasoning-conditioned retrieval.

The retrieved context is shared across reasoning steps rather than pre-segmented; instead, the reasoning plan governs how evidence is interpreted and composed during generation. This design decouples what evidence is retrieved from how it is used, enabling fine-grained analysis of evidence grounding and reasoning faithfulness.

\section{ArtCoT-QA: A benchmark for multimodal artwork reasoning}
To evaluate multimodal artwork understanding beyond final-answer correctness, we introduce \textbf{ArtCoT-QA}, a benchmark designed for analyzing multi-step reasoning and evidence grounding in complex artwork questions. ArtCoT-QA focuses on interpretive queries that require coordinated reasoning over visual content, artwork metadata, and external contextual knowledge, reflecting the structured reasoning practices common in art interpretation.
Unlike existing datasets~\cite{semart_2018, 2019_Artpedia} that primarily assess answer quality, ArtCoT-QA is designed as a \emph{diagnostic} benchmark. It provides explicit annotations of reasoning steps and evidence requirements, enabling fine-grained analysis of how reasoning, retrieval, and generation interact in multimodal systems. This makes the benchmark particularly suitable for evaluating agentic and retrieval-augmented approaches that structure the reasoning process.

\subsection{Motivation and design objectives}
Fine-grained artwork understanding involves an ordered reasoning process, such as identifying visual elements, linking them to contextual knowledge, and synthesizing interpretive conclusions. However, existing art-related datasets either provide only final explanations or focus on short factual answers, limiting the evaluation of reasoning coherence and evidence grounding.
ArtCoT-QA addresses this gap by making reasoning structure and grounding explicit. The benchmark is guided by three objectives:  
(1) \textbf{Multi-step reasoning}, where each question requires decomposition into multiple ordered reasoning steps;  
(2) \textbf{Explicit evidence grounding}, where each reasoning step is associated with a primary evidence source (e.g., visual content, metadata, or structured background knowledge); and  
(3) \textbf{Retrieval-faithful evaluation}, analyzing whether retrieved context correctly supports each reasoning step.

\vspace{-2mm}
\begin{table}[ht]
\caption{Statistics of the ArtCoT-QA dataset.}
\vspace{-3mm}
\label{tab:artcot-qa-stats}
\centering
\small
\setlength{\tabcolsep}{4pt}
\begin{tabular}{l r}
\toprule
\textbf{Statistic} & \textbf{Value} \\
\midrule
Questions / Paintings & 227 / 227 \\
Difficulty (High / Medium) & 174 / 53 \\
Multi-hop questions & 227 (100\%) \\
Reasoning  steps (min / max / mean) & 4 / 5 / 4.7 \\
Question length (min / max / mean) & 15 / 38 / 23 \\
Answer length (min / max / mean) & 95 / 186 / 138 \\
\bottomrule
\end{tabular}
\vspace{-4mm}
\end{table}

\subsection{Dataset construction and annotation}

ArtCoT-QA is built from curated subset of the SemArt dataset, a multimodal art dataset that provides painting images, structured metadata (title, author, technique, timeframe, tags), and textual descriptions organized by perspective (content, form, context). We use a \emph{filtered subset} of SemArt so that each selected painting satisfies: (i)~\emph{rich metadata} non-empty meta data---and (ii)~\emph{informative descriptions}, sufficient length (word count above a threshold), presence of semantic keywords relevant to reasoning (symbolism, depiction, context, meaning), and at least both \emph{content} and \emph{context} perspectives. This favors paintings that support non-trivial, multi-step questions rather than simple factual lookups. 

For each selected painting, an SOTA MLLM (Anthropic Claude-4.5 Sonnet \cite{claude-sonnet-4.5}) is used as a controlled annotation tool to structure supervision from existing visual and textual sources. The model receives the painting image, metadata, and formatted description, and is prompted to produce a single question–answer pair with an explicit chain-of-thought (CoT) and grounding tags. The prompt enforces: (1)~questions that require multi-step reasoning (no direct factual questions answerable from metadata alone); (2)~answers fully supported by the provided metadata, description, and visible content; (3)~CoT steps with exactly one grounding tag per step from the fixed schema (Visual, Metadata, KG-Background, Common-Knowledge); (4)~difficulty and planning-complexity labels consistent with the number of steps and evidence types; and (5)~diverse question phrasings (e.g., ``What visual elements\ldots'', ``In what way\ldots'', ``What evidence\ldots'') to avoid template-like wording. Outputs are parsed and validated as JSON; invalid or off-spec responses are discarded. 
Automated checks using GPT-5.2\cite{GPT-5-2} were additionally employed to flag potential inconsistencies and annotation errors. All construction and annotation steps were followed by a manual validation stage conducted by a researcher with domain expertise in visual art understanding. All constructions details are in the codebase for reproducibility and further research.

\paragraph{Resulting dataset.} As shown in Table~\ref{tab:artcot-qa-stats}, the released ArtCoT-QA set used in this work comprises one question over each of the \textbf{227 unique paintings}, with a spread of difficulty (e.g., high and medium in the current slice), planning complexity (multi-hop by design for the main slice), and CoT length (e.g., 4--5 steps per question on average). Each item includes the question, the reference answer, the CoT steps with grounding tags, the set of evidence types used, difficulty and planning-complexity labels, painting identifier, image path, metadata, and the SemArt description.

\begin{table*}[t]
\centering
\caption{Overall artwork explanation performance on Artpedia and SemArt. A-MAR consistently improves over static ArtRAG across both Claude-4.5-Haiku and Mistral-large-3 backbones, demonstrating that the gains from agent-based query planning generalize across multimodal language models. We use ``--'' to indicate values missing in the original works.}
\vspace{-2mm}
\begin{tabular}{llcccccccc}
\toprule
\textbf{Dataset} & \textbf{Models} & \textbf{BLEU-1} & \textbf{BLEU-2} & \textbf{BLEU-3} & \textbf{BLEU-4} & \textbf{METEOR} & \textbf{SPICE} & \textbf{ROUGE-L} & \textbf{CLIP} \\
\midrule
\multirow{7}{*}{Artpedia} 
 & Wu 2022~\cite{Wu_2022}  
 & 24.7 & -- & -- & 3.06 & 6.58 & -- & 22.4 & -- \\
 & KALE (w/o metadata)~\cite{ijcai2024p848}  
 & 29.9 & 15.0 & 7.95 & 4.77 & 8.02 & 5.49 & 22.4 & -- \\
 & KALE (w/ metadata)~\cite{ijcai2024p848}   
 & 32.6 & 17.7 & 10.9 & 7.48 & 9.33 & 7.68 & 23.7 & -- \\
 & ArtRAG~\cite{ArtRAG_MM2025} (Mistral-large-3)     
 & 39.2 & 19.2 & 13.1 & 11.2 & 14.1 & 10.4 & 27.5 & 84.6 \\
 & ArtRAG~\cite{ArtRAG_MM2025}  (Claude-4.5-Haiku)    
 & 39.4 & 19.5 & 12.9 & 10.7 & 14.0 & 10.2 & 27.3 & 84.9 \\
 \rowcolor{gray!20} 
 & \textbf{A-MAR (Mistral-large-3)}   
 & \textbf{43.3} & 22.2 & 15.2 & \textbf{12.4} & \textbf{16.3} & \textbf{12.3} & \textbf{29.4} & 86.4 \\
 \rowcolor{gray!20} 
 & \textbf{A-MAR (Claude-4.5-Haiku)}   
 & 43.2 & \textbf{22.4} & \textbf{15.8} & \textbf{12.4} & 16.1 & 12.2 & 29.3 & \textbf{86.5} \\
\midrule
\multirow{7}{*}{SemArt} 
 & Bai~\cite{Bai2021ExplainMT}   
 & -- & -- & -- & 9.1 & 11.4 & -- & 23.1 & -- \\
 & KALE (w/o metadata)~\cite{ijcai2024p848}   
 & 25.9 & 13.8 & 8.8 & 6.7 & 7.5 & 6.0 & 19.9 & -- \\
 & KALE (w/ metadata)~\cite{ijcai2024p848}   
 & 27.7 & 15.7 & 10.8 & 8.6 & 9.5 & 7.3 & 21.9 & -- \\
 & ArtRAG~\cite{ArtRAG_MM2025} (Mistral-large-3)     
 & 30.3 & 15.4 & 10.2 & 5.2 & 10.2 & 9.2 & 24.2 & 84.7 \\
 & ArtRAG~\cite{ArtRAG_MM2025} (Claude-4.5-Haiku)     
 & 30.1 & 15.2 & 10.2 & 4.6 & 10.1 & 9.1 & 24.2 & 84.2 \\
 \rowcolor{gray!20} 
 & \textbf{A-MAR (Mistral-large-3)}   
 & 33.7 & \textbf{17.2} & 11.3 & \textbf{5.3} & 12.3 & \textbf{9.3} & 26.3 & \textbf{86.3} \\
 \rowcolor{gray!20} 
 & \textbf{A-MAR (Claude-4.5-Haiku)}   
 & \textbf{34.2} & 16.1 & \textbf{12.6} & 5.2 & \textbf{12.6} & 9.6 & \textbf{26.5} & 86.2 \\
\bottomrule
\end{tabular}

\label{tab:overall_performance}
\end{table*}

\begin{table*}[t]
\centering
\caption{ ArtCoT-QA evaluation on retrieval grounding and multi-step reasoning under the same backbone model and retrieval budget. We compare a CoT-only baseline without retrieval, static retrieval, a text-only planner, and A-MAR. Step Completeness and Faithfulness measure \emph{reasoning ability}, Subgraph Relevance and Evidence Coverage assess \emph{retrieval grounding} (``--'' indicates not applicable), and Answer Quality measures final response quality. Boldface denotes the best result per backbone.}
\vspace{-2mm}

\small
\begin{tabular}{llccccc}
\toprule
\textbf{Model} 
& \textbf{Variant} 
& \textbf{Step Completeness} 
& \textbf{Faithfulness}  
& \textbf{Subgraph revelance} 
& \textbf{Evidence Coverage}  
& \textbf{Answer Quality} \\
\midrule
\multirow{4}{*}{Claude-4.5 Haiku} 
& MLLM-CoT
& 3.23 & 3.12 & -- & -- & 2.86 \\
& Static retrieval
& -- & -- & 2.37 & 2.16	& 3.12 \\
& Text-only Planner 
& -- & --  & 2.78 & 2.72	& 3.33 \\
\rowcolor{gray!15}
& \textbf{A-MAR} 
& \textbf{3.58} & \textbf{3.45} & \textbf{2.88} & \textbf{2.82} & \textbf{3.63} \\
\midrule
\multirow{4}{*}{Mistral-3 Large} 
& MLLM-CoT 
& 3.12 & 3.01 & -- & -- & 2.61 \\
& Static retrieval
& -- & -- & 2.53 & 2.31	& 2.66 \\
& Text-only Planner
& -- & -- & 2.72 &	2.66 &	3.36 \\
\rowcolor{gray!15}
& \textbf{A-MAR} 
& \textbf{3.67} & \textbf{3.44} & \textbf{2.85} & \textbf{2.78} & \textbf{3.65} \\
\bottomrule
\end{tabular}
\label{tab:artcot_results}
\end{table*}

\subsection{Reasoning, grounding, and evaluation}

ArtCoT-QA adopts a step-level reasoning and grounding schema to enable fine-grained analysis of reasoning and retrieval behavior. Each question is annotated with an ordered sequence of reasoning steps, and each step is associated with exactly one primary grounding category: \textit{Visual}, \textit{Metadata}, \textit{Description}, \textit{KG-Background}, or \textit{Common-Knowledge}.
Each reasoning step corresponds to a distinct \emph{evidence requirement}.  Enforcing a single primary grounding source per step avoids ambiguous attribution and enables precise analysis of whether systems retrieve and utilize appropriate evidence at the correct stage of reasoning. This structured annotation directly supports the evaluation of reasoning completeness, grounding fidelity, and retrieval alignment reported in Section~\ref{sec:eval_metrics}.

\section{Experiments and Results}

We conducted extensive experiments both quantitatively and qualitatively to evaluate the effectiveness of A-MAR, focusing on the following key research questions:

\begin{itemize}
  \item RQ-1: Does agent-based multimodal retrieval improve overall artwork explanation quality?
  \item RQ-2: How does agent-based multimodal retrieval improve retrieval grounding and multi-step reasoning?
  \item RQ-3: How does reasoning-aware, multimodal query planning improve fine-grained artwork understanding?
  \item RQ-4: What qualitative analysis says about reasoning and retrieval behavior of A-MAR?
\end{itemize}

\vspace{-3mm}
\subsection{Experimental setup}

\subsubsection{Models and baselines}

To ensure fair and controlled comparisons, we fix the MLLM backbone across all RAG-based methods unless otherwise stated. Our primary backbone is Anthropic Claude~4.5~Haiku~\cite{claude-haiku-4.5}, a relatively compact and strong multimodal model, selected for its reliable structured generation and suitability for controlled ablation studies. To test robustness across model families and architectures, we additionally report selected results using Mistral-3 Large~\cite{Mistral-Large-3-675b}, a recent Mixture-of-Experts (MoE) model representing a popular and distinct architectural paradigm. This allows us to verify that the observed gains are not specific to a single model family, scale, or architecture.
We compare A-MAR against the following baselines:
\emph{MLLM-CoT:} direct generation with chain-of-thought reasoning but without external retrieval;
\emph{Static Retrieval (No Planner):} structured retrieval over the Art Context Knowledge Graph (ACKG)\cite{ArtRAG_MM2025} without explicit planning;
\emph{Text-only Planner:} generates a retrieval intent using only textual inputs (question and metadata), without visual grounding;
\emph{A-MAR:} employs a multimodal planner that decomposes the query into explicit reasoning steps and evidence requirements to guide retrieval.
We further compare with prior art-specific methods, including Wu 2022~\cite{Wu_2022}, Bai~\cite{Bai2021ExplainMT}, KALE~\cite{ijcai2024p848}, and ArtRAG~\cite{ArtRAG_MM2025}. A-MAR is agnostic to the underlying external knowledge source; in this work, all retrieval-based methods use the same structured Art Context Knowledge Graph (ACKG)~\cite{ArtRAG_MM2025} to ensure controlled experimentation and fair comparison. In all retrieval based methods, we control the same retrieval budget, with the multi-granularity structured context retriever selects $k = 10$ nodes at the coarse level and $m = 5$ nodes at the fine-grained level. We use a softmax function for score normalization, and set the weighting score $\lambda = 0.5$. All MLLM inferences use default decoding parameters (e.g., temperature, sampling settings) unless otherwise specified.

\begin{figure*}[]
\vspace{-3mm}
    \centering\includegraphics[width=\linewidth]{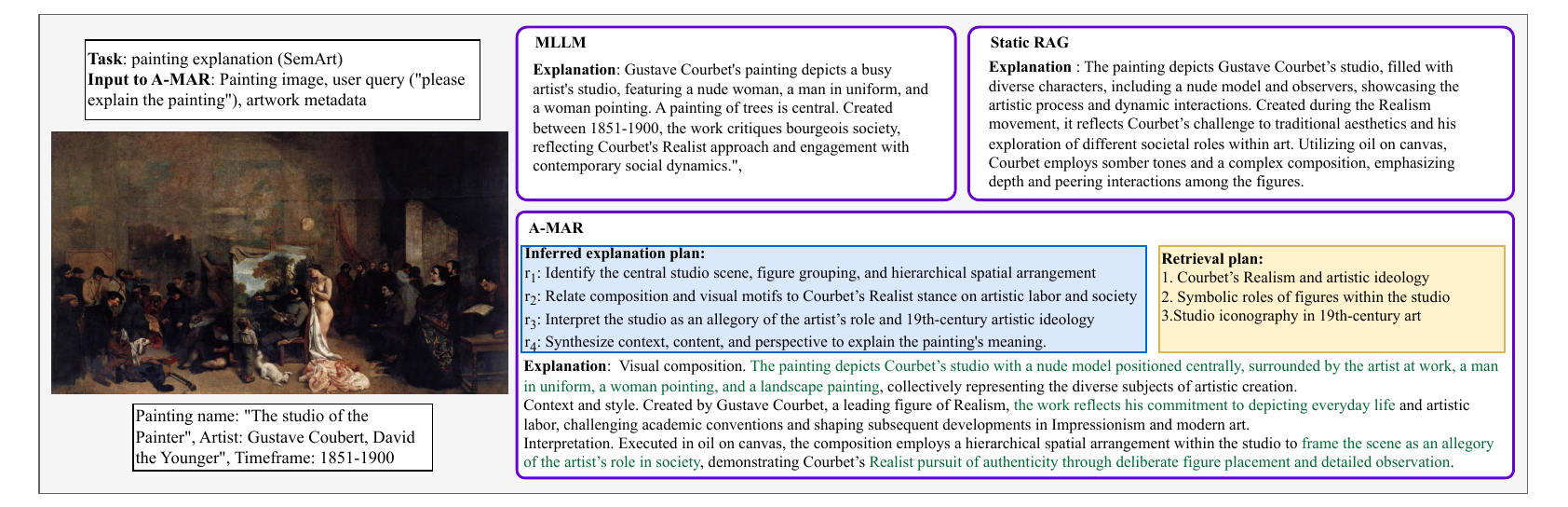}
    \vspace{-6mm}
    \caption{A-MAR autonomously plans how to explain a painting under an open-ended SemArt prompt, retrieves targeted contextual knowledge, and produces a structured, art-historically grounded explanation that static RAG fails to achieve.}
    \label{fig:qualitative_semart}
    \vspace{-2mm}
\end{figure*}

\subsubsection{Datasets} 
We evaluate our approach on three datasets, each serving a distinct experimental purpose. SemArt~\cite{semart_2018} and Artpedia~\cite{2019_Artpedia} are used to assess overall artwork explanation quality and maintain comparability with prior work. These datasets contain expert-written or curated explanations spanning visual content, artistic style, and contextual background. ArtCot-QA, introduced in this work, is designed specifically for reasoning-intensive evaluation. It consists of interpretive artwork questions that require multi-step reasoning across multiple evidence modalities, paired with annotated reasoning steps and step level grounding labels.

\begin{table*}[t]
\centering
\caption{Ablation study on the query planner for Agentic ArtRAG. We fix the generation and reasoning components and vary only the retrieval planning strategy. Bold indicates the best result per dataset.}
\vspace{-2mm}
\small
\begin{tabular}{llccccccc}
\toprule
\textbf{Dataset} & \textbf{Model Variant} & \textbf{BLEU-1} & \textbf{BLEU-2} & \textbf{BLEU-3} & \textbf{METEOR} & \textbf{SPICE} & \textbf{ROUGE-L} & \textbf{CLIP} \\
\midrule
\multirow{4}{*}{Artpedia} 
& (a) Static retrieval            
& 40.2 & 19.8 & 12.5 & 14.2 & 10.7 & 27.8 & 85.2 \\
& (b) + Text-only Planner 
& 42.3 & 20.6 & 14.7 & 15.6 & 11.5 & 28.6 & 85.7 \\

& \cellcolor{gray!20}(c) + Full Planner (Ours)  
& \cellcolor{gray!20} \textbf{43.2} & \cellcolor{gray!20}\textbf{22.4} & \cellcolor{gray!20}\textbf{15.8} & \cellcolor{gray!20}\textbf{16.1} & \cellcolor{gray!20}\textbf{12.2} & \cellcolor{gray!20}\textbf{29.3} & \cellcolor{gray!20}\textbf{86.5} \\
\midrule
\multirow{4}{*}{SemArt} 
& (a) Static retrieval             
& 31.2 & 15.2 & 10.5 & 10.4 & 9.15 & 24.8 & 85.1 \\
& (b) + Text-only Planner    
& 32.5 & 15.4 & 11.0 & 10.8 & 9.34 & 24.91 & 85.6 \\
& \cellcolor{gray!20}(c) + Full Planner (Ours)  
& \cellcolor{gray!20} \textbf{34.2} & \cellcolor{gray!20}\textbf{16.1} & \cellcolor{gray!20}\textbf{12.6} & \cellcolor{gray!20}\textbf{12.6} & \cellcolor{gray!20}\textbf{9.6} & \cellcolor{gray!20}\textbf{26.3} & \cellcolor{gray!20}\textbf{86.2} \\
\bottomrule
\end{tabular}
\label{tab:ablation}
\vspace{-3mm}
\end{table*}

\subsubsection{Evaluation Metrics}
\label{sec:eval_metrics}
We adopt different evaluation protocols depending on the target capability assessed. For SemArt and Artpedia, we follow prior work and report standard automatic metrics for artwork explanation generation, including BLEU-4~\cite{2002-bleu}, METEOR~\cite{2005-meteor}, SPICE~\cite{spice_2016}, ROUGE-L~\cite{2004_rouge}, and CLIPScore~\cite{2021-clipscore}.
However, surface-level overlap metrics are insufficient for evaluating multi-step, open-ended reasoning with explicit evidence grounding, as required by ArtCoT-QA. They cannot assess the correctness or ordering of intermediate reasoning steps, nor whether retrieved evidence is appropriately selected and used.
To address these limitations, we adopt an LLM-as-a-Judge evaluation protocol, which correlates well with expert human judgments while remaining scalable for large open-vocabulary benchmarks~\cite{chen2024mllm,gu2024survey,liu-etal-2023-g, huang2025image2text2image}. This paradigm is well suitable for ArtCoT-QA, where multiple reasoning trajectories may be valid and evaluation must jointly consider step-level faithfulness, grounding correctness, and retrieval relevance.
The LLM is used solely as a consistent scoring tool, similar with human annotators. The judge evaluates model outputs with groundtruth along three dimensions on a 1–5 scale: Answer Quality, Reasoning Quality, and Retrieval Quality. We use Claude 4.5 Sonnet~\cite{claude-sonnet-4.5} as the judge model, which is different than all generation models used in our experiments, mitigating self-evaluation bias.

\vspace{-1mm}
\subsection{Overall performance}

Table~\ref{tab:overall_performance} reports results on Artpedia and SemArt. Across all metrics and both backbones (Mistral-3 Large and Claude-4.5 Haiku), A-MAR consistently outperforms ArtRAG and prior art-specific baselines. On Artpedia, A-MAR (Mistral-3) achieves the best performance across BLEU-1/2/3/4 (43.3 / 22.2 / 15.2 / 12.4), METEOR (16.3), SPICE (12.3), and ROUGE-L (29.4), surpassing both KALE variants (with and without metadata) as well as static ArtRAG.
Similar gains are observed on SemArt, where A-MAR improves BLEU-4 to 5.3 and achieves strong semantic scores (METEOR = 12.3, SPICE = 9.3, ROUGE-L = 26.3). The improvements are consistent across model architectures, indicating that the performance gains stem from reasoning query planning rather than specific backbone characteristics. Overall, these results provide a clear positive answer to RQ1, demonstrating that reasoning-conditioned retrieval improves overall artwork explanation quality.

\begin{figure*}[ht]
    \centering\includegraphics[width=\linewidth]{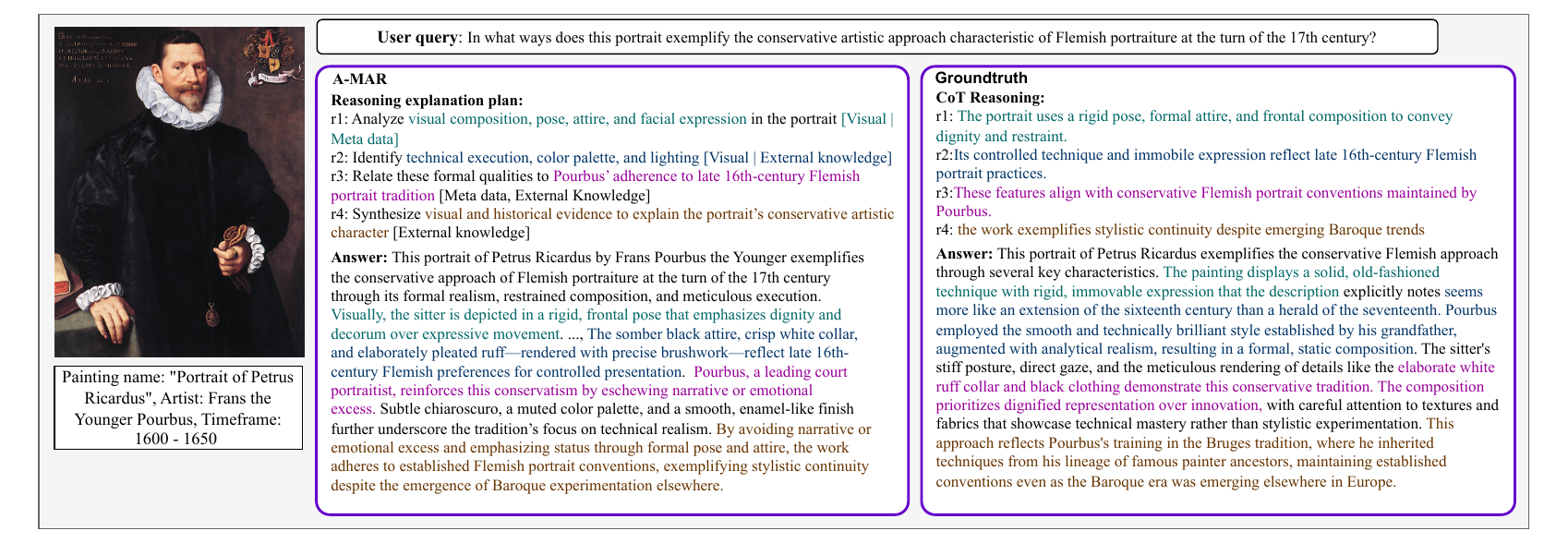}
    \vspace{-7mm}
    \caption{ArtCoT-VQA qualitative example illustrating step-level reasoning alignment between A-MAR and ground-truth reasonings. Given a knowledge-intensive art user query, A-MAR infers a structured explanation plan and generates step-wise reasoning grounded in visual evidence and art-historical context, matching the groundtruth.(indicated by matching colors)}
    \label{fig:qualitative_artcot}
    \vspace{-2mm}
\end{figure*}

\subsection{Reasoning-conditioned retrieval analysis}
\label{sec:Reasoning-Conditioned_exp}

We now directly evaluate whether agent-based planning enables \emph{reasoning-conditioned retrieval}, which constitutes the core claim of A-MAR. Using ArtCoT-QA, we compare \textbf{MLLM-CoT}, \textbf{Static Retrieval}, a \textbf{Text-only Planner} that performs retrieval planning without explicit reasoning decomposition, and \textbf{A-MAR} with full agent-based multimodal planning, under the same backbone model and retrieval budget.
Table~\ref{tab:artcot_results} reports performance on reasoning and retrieval metrics. Introducing text-only retrieval planning improves retrieval-oriented measures such as subgraph relevance and evidence coverage compared to Static Retrieval. However, these gains translate only marginally into answer quality and do not yield faithful multi-step reasoning, indicating that improved access to relevant evidence alone is insufficient for complex artwork questions.
In contrast, A-MAR, where retrieval is explicitly conditioned on a structured reasoning plan, achieves substantial improvements in both step completeness and reasoning faithfulness across backbones. These gains are accompanied by the highest answer quality scores, demonstrating that evidence must be retrieved \emph{for the appropriate reasoning step}, rather than merely retrieved accurately.
Notably, the Text-only Planner achieves retrieval quality comparable to A-MAR but lags significantly in reasoning metrics. This gap highlights the necessity of explicit reasoning decomposition: retrieval control alone cannot ensure grounded multi-step reasoning without a plan that specifies how retrieved evidence is to be used.

\subsection{Ablation studies on retrieval query planning}
To address RQ-3 — how reasoning-aware, multimodal query planning improves fine-grained artwork understanding — we conduct a controlled ablation study that isolates the effect of the retrieval planning strategy in A-MAR. In this experiment, we fix the backbone MLLM, generation procedure, reasoning format, and external knowledge source, and vary only how retrieval queries are planned.
As shown in Table~\ref{tab:ablation}, introducing an explicit planner consistently improves performance over static retrieval across both Artpedia and SemArt, demonstrating that reasoning-conditioned query formulation alone already benefits retrieval-augmented generation. More importantly, the full multimodal planner yields the strongest gains across all evaluation metrics, including lexical overlap (BLEU, ROUGE-L), semantic alignment (METEOR, SPICE), and vision–language consistency (CLIP).
 Overall, this ablation confirms that query planning enables the retriever to surface evidence that is better aligned with the model’s multi-step reasoning needs.

\subsection{Qualitative examples}

Figures~\ref{fig:qualitative_semart} and~\ref{fig:qualitative_artcot} provide qualitative examples illustrating how A-MAR enables reasoning-conditioned retrieval and structured explanation generation.
Figure~\ref{fig:qualitative_semart} presents an open-ended painting explanation example from SemArt. Compared to a standalone MLLM and static RAG, A-MAR first infers an explicit explanation plan that decomposes the task into semantically meaningful steps, including visual analysis, stylistic interpretation, and historical contextualization. Guided by this plan, A-MAR retrieves targeted contextual knowledge such as Courbet’s Realist ideology and symbolic roles, and integrates it coherently into the final explanation. In contrast, static RAG retrieves relevant but unstructured information, resulting in a descriptive yet loosely organized explanation that lacks clear interpretive progression.
Figure~\ref{fig:qualitative_artcot} shows an ArtCoT-QA example that highlights step-level reasoning alignment. Given a knowledge intensive query about conservative Flemish portraiture, A-MAR generates a structured reasoning plan that closely mirrors the annotated ground truth reasoning steps. Each reasoning step is grounded in either visual evidence (e.g., pose, attire, composition) or retrieved art-historical context (e.g., Flemish portrait conventions), demonstrating faithful coordination between reasoning and retrieval. The resulting explanation exhibits strong correspondence with the ground truth CoT, both in content and ordering.

\vspace{-3mm}
\section{Discussion and conclusion}

In this work we present A-MAR, an agent based retrieval augmented framework that explicitly separates reasoning, retrieval, and generation for multimodal artwork understanding. By externalizing reasoning as a structured plan and conditioning retrieval on reasoning intent, A-MAR enables interpretable, evidence-grounded explanations and supports fine grained evaluation of reasoning faithfulness and retrieval quality. To facilitate such analysis, we introduce ArtCoT-QA, a diagnostic benchmark that exposes step level reasoning structure and evidence requirements, moving beyond final answer evaluation in artwork question answering.
While A-MAR consistently outperforms static retrieval baselines, the specific design choices governing agent behavior—such as how reasoning is decomposed, ordered, and expressed—are not unique. Multiple plausible reasoning formulations exist for artwork interpretation, including comparative, contrastive, analogical, or theme- versus form-driven strategies. An important direction for future work is to systematically investigate how different reasoning styles affect performance, and to develop  evaluation protocols for more controllable and transparent agent-based multimodal AI systems.

\section{Acknowledge}
This project received funding from University of Amsterdam Data Science for PhD research. This work used the Dutch national e-infrastructure with the support of the SURF Cooperative using grant No.EINF-14160.

\bibliographystyle{ACM-Reference-Format}
\bibliography{references}

\appendix

\end{document}


\title{Supplementary Materials for ArtRAG: A Retrieval-Augmented Framework for Visual Art Understanding with Structured Context }

\settopmatter{printacmref=false}

\maketitle

\tableofcontents

Due to space constraints in the main paper, we omit certain implementation details—such as full prompting strategies used for ACKG building—and limit some sections to a single illustrative example. In this supplementary material, we first provide the complete prompting details, followed by additional qualitative examples to offer a more comprehensive and convincing analysis of our method. \noindent Our code base is available at:
\url{https://anonymous.4open.science/r/ArtRAG_software-315A}

\begin{figure*}[h]
    \centering\includegraphics[width=0.9\linewidth]{images/graph_example.png}
    \vspace{-3mm}
    \caption{Example of automatic entity and relation extraction from Wikipedia text for Art Context Knowledge Graph (ACKG) construction. Starting from a scraped Wikipedia page (e.g., “Dutch art”), our framework extracts key entities such as Dutch golden age painting (Culture \& History), Dutch masters (Artist), and associated descriptions. It also identifies semantic relationships, such as “Dutch masters” being influential figures in the “Dutch golden age painting” era. The resulting nodes and edges are used to construct a structured graph capturing the interconnected context of art history.}
    \label{fig:graph_example}
\end{figure*}

\subsection{ACKG construction details}

We scraped over 1,000 Wikipedia pages shown in Tab.\ref{tab:wiki_stats} related to art, guided by category tags and entity types from SemArt, ArtPedia, and ExpArt. Each page was segmented into overlapping text chunks using a sliding window (1000 tokens with 100-token stride), enabling finer-grained concept and relation extraction. in Fig.\ref{fig:graph_example}, we show a example of A subgraph snapshot built from Wikipedia page "Dutch art".

Fuzzy Matching for Node Deduplication:
To reduce redundancy in the constructed graph, we perform post-processing to merge nodes with similar surface forms or semantically equivalent meanings. We compute the pairwise string similarity between node names using normalized Levenshtein distance, implemented via the RapidFuzz library. Two nodes are merged if their similarity exceeds a threshold of 0.95 and their types are compatible (e.g., both are "Art Movement" nodes).

In below, we show the prompt we used to extract nodes and edges from wikipedia documents:

\begin{table}[h]
    \centering
    \caption{Statistics of collected Wikipedia pages and word counts for different art-related categories.}
    \begin{tabular}{lccc}
        \toprule
         \textbf{Document Type} & \textbf{ Pages} & \textbf{Word number} & \textbf{Token number}\\
        \midrule
        Artists         & 952     & 1750K   & 2067K\\
        Art schools      & 20      & 69K     & 80K \\
        Art Types        & 9       & 37K     &  43K \\
        Cultural events & 85      & 781K    &  895K\\
        Art movements   & 25      & 128K    &  159K  \\
        \midrule
        Total           & 1091    & 2765K  & 3244K\\
        \bottomrule
    \end{tabular}

    \label{tab:wiki_stats}
\end{table}

\begin{tcolorbox}[breakable, fontupper=\customfont, title=Prompt for Node and edge extraction]
{\small
\textbf{Input Data}: Art related Document chunk\\
\textbf{Instruction}: 
Given a visual art related text document that is potentially relevant to this activity and a list of entity types, identify all entities of those types from the text and all relationships among the identified entities.

\textbf{Steps}:
1. Identify all entities. For each identified entity, extract the following information: \\
- entity\_name: Name of the entity, capitalized \\
- entity\_type: One of the following types: [{entity\_types}] \\
- entity\_description: Comprehensive description of the entity's attributes and activities

2. From the entities identified in step 1, identify all pairs of (source\_entity, target\_entity) that are *clearly related* to each other. \\
For each pair of related entities, extract the following information: \\ 
- source\_entity: name of the source entity, as identified in step 1 \\ 
- target\_entity: name of the target entity, as identified in step 1 \\
- relationship\_description:

------ Example -----

\textit{Contextual Graph Snapshot:}
\begin{itemize}
  \item \textbf{Nodes:}....
  \item \textbf{Edges:}....
\end{itemize}

}
\end{tcolorbox}

\begin{table*}[h]
\centering 
\caption{Illustrative examples of node types and representative entities in the Art Context Knowledge Graph.}
\label{tab:node_examples}
\begin{tabular}{ll}
\toprule
\textbf{Node Type}         & \textbf{Example Entities} \\
\midrule
\textbf{Artist}                    & Paul Cézanne, Vincent van Gogh, Claude Monet, Jacques-Louis David \\
\textbf{Theme}                     & Working-class life, Spiritual devotion, Balance and Harmony \\
\textbf{Culture \& History}        & Flemish Renaissance, Dutch Golden Age, Christian faith  \\
\textbf{Art Style \& Technique}    & Oil on canvas, Fresco, Chiaroscuro, Linear perspective \\
\textbf{Art Movement \& School}    & Impressionism, Surrealism, Bauhaus, Florentine School \\
\bottomrule
\end{tabular}
\end{table*}

\begin{figure*}[h]
    \centering\includegraphics[width=0.9\linewidth]{images/Qual_example2_subg_SM.drawio.pdf}
    \vspace{-3mm}
    \caption{Visual example of the contextual subgraph $\mathcal{G}'$ retrieved by ArtRAG for painting \ptitle{Saint Ladislaus, King of Hungary.}}
    \label{fig:graph_example}
\end{figure*}

\subsection{Concept detection}
To support downstream context retrieval and graph grounding, we extract high-level visual and thematic concepts from each painting. We design a dedicated prompt that instructs a multimodal language model to analyze both the visual features of the painting and its associated metadata (e.g., title, artist, creation period) to generate a short list of relevant concepts, as shown bwlow:
\begin{tcolorbox}[breakable, fontupper=\customfont, title=Prompt for concept detection]
{\small
\textbf{Input Data}:  Painting attributes, and visual painting \\
\textbf{Instruction}: 
You are an expert in art interpretation. Given an image of a painting along with its title, artist, and creation period, identify a list of high-level concepts that capture the key concptes of the painting.

\textbf{\color{black}{Formatting and Constraints}}: 
Output should be formatted in Markdown. If information is missing, do not hallucinate or fabricate details.
}
\end{tcolorbox}

\begin{figure*}
\centering\includegraphics[width=0.9\linewidth]{images/Human_eval_example.drawio-2.pdf}
    \vspace{-3mm}
    \caption{ Human evaluation form design, containing the description of the task, the painting information, and three random shuffled result.}
    \label{fig:human_eval_example}
\end{figure*}

\subsection{Human evaluation details}
Figure \ref{fig:human_eval_example} shows the human evaluation interface used in our study. Each task instance presents annotators with a painting image, associated metadata (title, artist, time period), and three model generated descriptions, shown in randomized order. Participants are asked to assess and rank the explanations from best to worst based on criteria such as clarity, relevance, and contextual depth. The interface also includes detailed instructions to guide the annotators and ensure consistent evaluations across participants.

\begin{figure*}[ht]
    \centering\includegraphics[width=0.8\linewidth]{images/Qual_example_Conver2_SM.drawio.pdf}
    \caption{ Qualitative examples of ArtRAG answering multi-perspective questions about paintings. The painting “Wooded Landscape” by Jan Hackaert is interpreted from two contextual angles: personal beliefs and artistic style. ArtRAG integrates cultural and stylistic signals—such as Dutch Golden Age pastoral themes and landscape harmony}
    \label{fig:framework_b}
\end{figure*}

\begin{figure}[h]
\begin{tcolorbox}[breakable, fontupper=\customfont, title=Few-shot learning generation]
{\small
\textbf{Input Data}: Query, Painting attributes, and a contextual subgraph with nodes and edges. \\
\textbf{Instruction}: 
You are an expert in art interpretation. Given visual and contextual information about a painting, your task is to generate a concise and structured explanation.

\textbf{\color{black}{Formatting and Constraints}}: 
If information is missing, do not hallucinate or fabricate details.

------------------------ Example -----------------------

\textit{Painting Metadata:} .....\\
\textit{Contextual Subgraph:}
Nodes:....,  Edges:.... \\
\textit{Generated Explanation:}.....

---------------------------------------------------------------

\textbf{Test Input:} \{Query\}, \{Painting Metadata\}, \{Subgraph $\mathcal{G}'$\} \\
\textbf{Expected Output:} Painting explanation

}
\end{tcolorbox}
\vspace{-2mm}
\caption{Prompt template used for augmented generation with in-context
learning}
\vspace{-2mm}
\label{fig:prompt_gen}
\end{figure}

\subsection{More qualitative examples}
We provide additional qualitative examples that showcase the effectiveness of ArtRAG in generating rich and context-aware explanations of visual artworks. 
In Each example includes the painting image, its associated metadata, and descriptions generated by three different models: ArtRAG, GPT-4o, and KALE. These examples illustrate how ArtRAG integrates visual and contextual knowledge to produce more informative, interpretable, and historically grounded outputs. 
In particular, we highlight passages that reflect nuanced understanding of artistic style, cultural significance, or historical references, demonstrating the benefits of structured context retrieval.

\begin{figure*}[ht]
    \centering\includegraphics[width=\linewidth]{images/Qual_example1_SM.drawio.pdf}
    \caption{ Qualitative comparison of painting explanations by ArtRAG and GPT-4o on religious-themed artworks. ArtRAG captures the symbolic and stylistic nuances of the Renaissance period, interpreting how religious scenes reflect broader humanistic and theological trends. For instance, in “Virgin and Child with Eight Angels”, ArtRAG identifies the emphasis on emotional expression, sacred iconography, and evolving artistic techniques in early Italian Renaissance art. Similarly, in “Holy Family at the Table”, ArtRAG contextualizes the scene within domestic religious themes and Renaissance values. Compared to GPT-4o, which provides factual summaries, ArtRAG demonstrates deeper interpretive reasoning and awareness of historical context.}
    \label{fig:framework_b}
\end{figure*}

\begin{figure*}[ht]
    \centering\includegraphics[width=\linewidth]{images/Qual_example2_SM.drawio.pdf}
    \caption{ Qualitative comparison of explanation generation between ArtRAG and GPT4o.For “Fête in a Wood” by Nicolas Lancret, ArtRAG highlights Rococo-era social customs, aesthetic values, and cultural influences, offering contextual richness beyond surface-level visual elements. In contrast, GPT4o focuses on scene composition without deeper historical framing. In “Portrait of John 20th Earl of Crawford and Lindsay”, ArtRAG identifies the significance of aristocratic representation in early 18th-century Scottish portraiture, connecting visual elements to socio-political context. GPT4o, however, produces a hallucinated interpretation unrelated to the painting, underscoring the importance of context-aware generation. }
    \label{fig:framework_b}
\end{figure*}


\begin{figure*}[ht]
    \centering\includegraphics[width=\linewidth]{images/Qual_example_Conver_SM.drawio.pdf}
    \caption{Human evaluation example illustrating ArtRAG’s ability to interpret paintings through historical and thematic lenses. For “The Houses of Parliament in London” by Claude Monet, ArtRAG captures how the painting reflects Impressionist ideals and cultural context—highlighting the foggy urban landscapes of 19th-century London and the artist's engagement with light and atmosphere. It also conveys the philosophical themes of transience and fleeting perception, linking visual elements to Monet’s broader artistic intent and the Impressionist movement’s ethos.}
    \label{fig:framework_b}
\end{figure*}

\begin{figure*}[h]
    \centering\includegraphics[width=\linewidth]{images/Qual_example_Conver3_SM.drawio.pdf}
    \caption{ Qualitative example highlighting ArtRAG’s ability to explain stylistic and thematic transitions. For “Virgin and Child with Eight Angels” by Tommaso del Mazza, ArtRAG connects religious symbolism, stylistic markers like gold background, and emotional tone to contextual shifts from medieval to early Renaissance art. These answers reflect a deep understanding of period-specific artistic values and personal devotional expression.}
    \label{fig:framework_b}
\end{figure*}

\bibliography{references}


\appendix